\documentclass[letterpaper]{article} % DO NOT CHANGE THIS
\usepackage{aaai24}  % DO NOT CHANGE THIS
\usepackage{times}  % DO NOT CHANGE THIS
\usepackage{helvet}  % DO NOT CHANGE THIS
\usepackage{courier}  % DO NOT CHANGE THIS
\usepackage[hyphens]{url}  % DO NOT CHANGE THIS
\usepackage{graphicx} % DO NOT CHANGE THIS
\usepackage{natbib}  % DO NOT CHANGE THIS AND DO NOT ADD ANY OPTIONS TO IT
\usepackage{caption} % DO NOT CHANGE THIS AND DO NOT ADD ANY OPTIONS TO IT
\usepackage{algorithm}
\usepackage{algorithmic}

\usepackage{booktabs}
\usepackage{pifont}
\usepackage{amssymb}
\usepackage{xcolor}

\usepackage{newfloat}
\usepackage{listings}
\DeclareCaptionStyle{ruled}{labelfont=normalfont,labelsep=colon,strut=off} % DO NOT CHANGE THIS
\floatstyle{ruled}
\newfloat{listing}{tb}{lst}{}
\floatname{listing}{Listing}
\title{AGENT: An Aerial Vehicle Generation and Design Tool \\ Using Large Language Models}
\author{
    Colin Samplawski,
    Adam D. Cobb,
    Susmit Jha
}
\affiliations{
    Computer Science Laboratory, SRI International
    colin.samplawski@sri.com, adam.cobb@sri.com, susmit.jha@sri.com
}

\nocopyright 
\usepackage{bibentry}
\begin{document}

\maketitle

\begin{abstract}
Computer-aided design (CAD) is a promising application area for emerging artificial intelligence methods. Traditional workflows for cyberphysical systems create detailed digital models which can be evaluated by physics simulators in order to narrow the search space before creating physical prototypes. A major bottleneck of this approach is that the simulators are often computationally expensive and slow. Recent advancements in AI methods offer the possibility to accelerate these pipelines. We  use  the recently released AircraftVerse dataset, which  is especially suited for developing and evaluating large language models for designs. AircraftVerse contains a diverse set of UAV designs represented via textual design trees together with detailed physics simulation results. Following the recent success of large language models (LLMs), we propose AGENT (Aircraft GENeraTor). AGENT is a comprehensive design tool built on the CodeT5+ LLM which learns powerful representations of aircraft textual designs directly from JSON files. We develop a curriculum of training tasks which imbues a single model with a suite of useful features. AGENT is able to generate designs conditioned on properties of flight dynamics (hover time, maximum speed, etc.). Additionally, AGENT can issue evaluations of designs allowing it to act as a surrogate model of the physics simulation that underlies the AircraftVerse dataset. We present a series of experiments which demonstrate our system's abilities. We are able to achieve strong performance using the smallest member of the CodeT5+ family (220M parameters). This allows for a flexible and powerful system which can be executed on a single GPU enabling a clear path toward future deployment.
\end{abstract}

\begin{figure}
    \centering    \includegraphics[width=0.9\columnwidth]{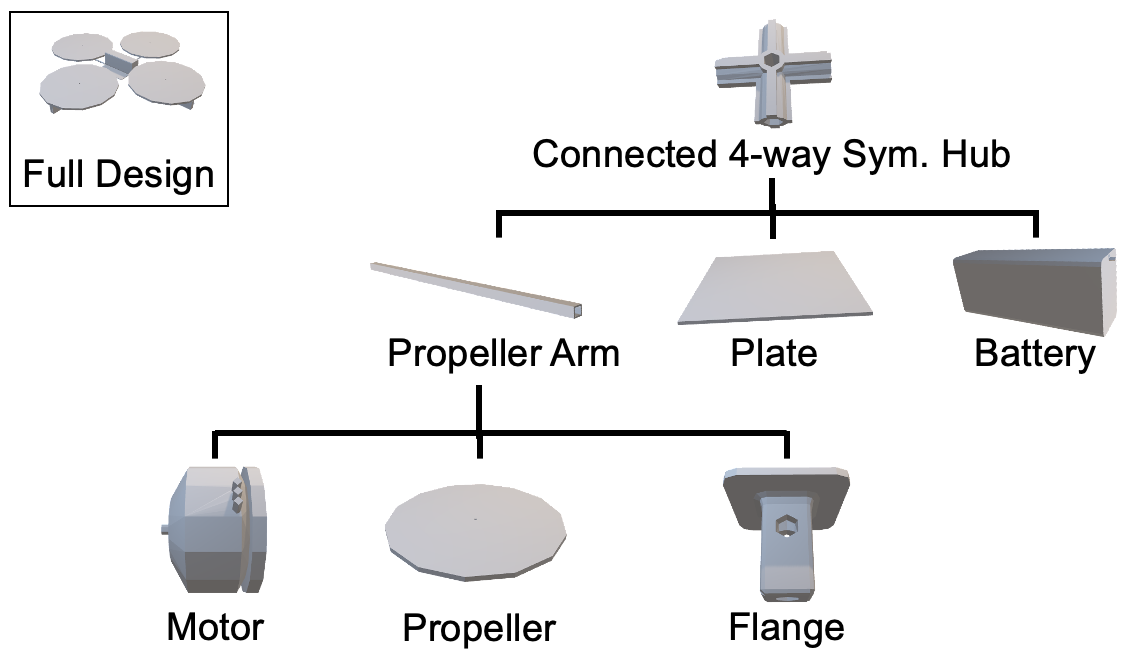}
    \caption{Graphical representation of a UAV JSON design tree from the AircraftVerse dataset.}
    \label{fig:design_tree}
\end{figure}

\section{Introduction}
Recent breakthroughs in generative modeling show the potential for AI to act as a design assistant in a number of domains. Well known examples include text-to-image models \cite{rombach2022high,dalle2}, music generation \cite{borsos2023audiolm}, and large language models \cite{bloom,llama}. It is natural to consider applying these ideas in engineering and scientific domains. Prior work has applied similar approaches to program generation \cite{codet5p,nijkamp2023codegen2}, pharmaceutical discovery \cite{korshunova2021openchem}, and mechanical component design \cite{granados2021machine,pai2022machine}. Unlike image or music generation, these domains act on highly structured data, where even small changes in the design can have large functional and performance changes as downstream effects. Furthermore, an AI assistant will need an implicit understanding of complex physical dynamics in order to generate reasonable designs. In this paper we consider such a domain, namely the design of unmanned aerial vehicles (UAVs). 

UAVs are becoming an increasingly mainstream technology with applications in many domains such as logistics \cite{gu2020vehicle}, agriculture \cite{aslan2022comprehensive}, search-and-rescue \cite{atif2021uav}, and construction \cite{jacob2021unmanned}. This diverse set of use cases calls for a high degree of diversity and specialization in UAV designs. The traditional design workflow for UAVs (and many other domains) consists of a human designer proposing designs which are then run through a simulator to receive feedback of the design's performance. These systems typically use advanced physics simulations which are costly and time consuming to compute. Afterwards the output of the simulation is parsed by the human operator, requiring continual human attention and often multiple runs of the simulator.  

Recently, the AircraftVerse dataset \cite{aircraftverse} was released to aid research in this domain. The AircraftVerse dataset contains 27,714 diverse UAV designs. Each design is represented by a design tree (Figure \ref{fig:design_tree}). The performance of each design was assessed using a simulation pipeline comprising the commercial tool Creo~\cite{creo} and a custom flight dynamics model ~\cite{walker2022flight,Bapty2022design}. Therefore each design is paired with the output of a physics simulator which includes flight dynamics performance such as hover time and flight distance. Both the designs and the simulator output are encoded as JSON files. 

AGENT (Aircraft GENeraTor) is built on top of the CodeT5+ large language model \cite{codet5p}. CodeT5+ is a state-of-the-art encoder-decoder transformer trained on a large dataset of publicly available code spanning many popular programming languages. We fine-tune CodeT5+ directly on the JSON files from the AircraftVerse dataset. We find that the model is able to quickly learn the tree structure of the dataset due to the similarity of JSON syntax to the programming languages it was trained on.

To fine-tune the CodeT5+ model we introduce a novel domain-specific instruction paradigm. During inference we supply a command (e.g. \texttt{<|generate\_design|>}) and any additional context (i.e. for conditioned generation) which instructs the decoder to output a relevant sequence. Using this paradigm, we are able to construct a curriculum of commands which allow a single model to exhibit a range of useful features. In addition to conditional design generation, we learn an inverse mapping (using an \texttt{<|evaluate\_design|>} command) which instructs the decoder to output the simulator results for a design in the encoder context. This allows the model to act as a surrogate for the underlying physics simulator. Finally, we train the model on a masking filling task (\texttt{fill\_masks}). This task enables a human operator to specify parts of a design and let AGENT fill the blanks with logical design choices. This capability is especially useful for generating designs from a fixed set of components. 

%The power of LLMs allow us to train a single model which supports all this behavior. We conduct a series of experiments to demonstrate the success of AGENT at these tasks. Finally, we discuss design workflows which are enabled by AGENT and path toward a full deployment system. (TODO)

The rest of this paper is structured as follows: we first give an overview of the AircraftVerse dataset and describe how we use it. We next describe our novel curriculum of training tasks. This leverages the power of LLMs to create a unified model with a suite of useful behaviors, which we validate in a series of experiments. We conclude by discussing the pathway to future deployment.

\begin{figure*}[ht]
    \centering    \includegraphics[width=0.9\textwidth]{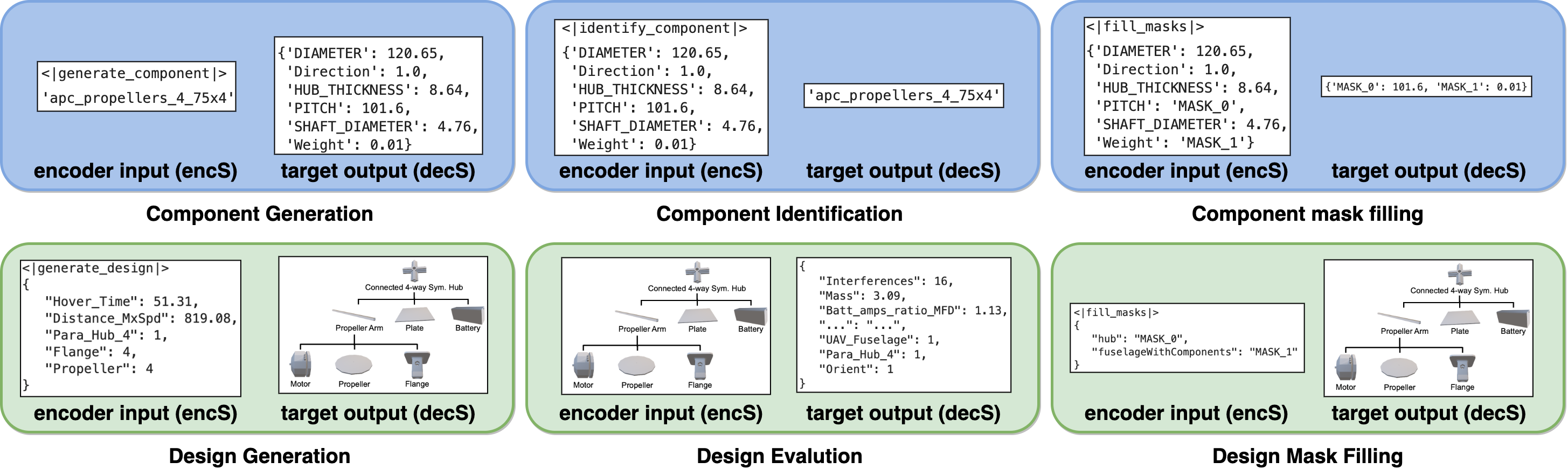}
    \caption{Curriculum of training tasks. %Note that design tree texts are shorted for the sake of space.
    }
    \label{fig:comp_tasks}
\end{figure*}

\section{Dataset}
In AircraftVerse, each design is specified by a design tree and is encoded in \texttt{design\_tree.json}. For each design, there is also an \texttt{output.json} file, containing the output data from the flight simulation pipeline. The \texttt{output.json} file is structured as a dictionary, including design metrics like \texttt{Hover\_Time} and \texttt{Max\_Distance}. As part of our approach, we also extend this dictionary to include counts of various components in the design, such as the number of propellers. These values can be directly inferred from the design tree. The dataset also includes a catalog of 727 vehicle components (e.g., motors) with their associated properties (e.g., mass). Each component is stored as a JSON dictionary. See Figure \ref{fig:comp_tasks} for an example.

\paragraph{Data preprocessing for CodeT5+.} We apply minimal preprocessing to the data before applying the CodeT5+ tokenizer. We use \texttt{json.dumps} from the standard Python package on the data dictionaries and then directly apply the tokenizer to this string. The design tree structure is encoded using the curly brace grammar of JSON. This format is familiar to the model, as it likely encountered similar structures during training from languages like Java and C. Moreover, sequences generated by a model trained on the JSON format can be easily converted directly into Python dictionaries.

As a final preprocessing step, we filter the overall dataset of designs such that those which are longer than 512 tokens are discarded due to the maximum sequence length of CodeT5+. We note that the designs with longer sequence lengths tended to be more complex and less likely to fly. Of the remaining designs, we then divide them into two classes: ``successful" designs and ``failed" designs. We define a design as successful if it has a hover time, maximum speed, and maximum distance that are all greater than 0. See Table \ref{table:data_stats} for a breakdown of the dataset statistics. For both the failed and successful designs, we create a random train/val/test split using 80\% of the designs for training, 10\% for validation, and 10\% for testing.

\begin{table}[h]
\centering
\begin{tabular}{|l|l|}
\hline
All Designs                        & 27,714 \\ 
Designs with \textless{}512 Tokens & 14,273 \\ 
Successful Designs                 & 5094   \\ 
Failed Designs                     & 9179   \\ 
Design Components                   & 727    \\ 
\hline
\end{tabular}
\caption{Dataset Statistics}
\label{table:data_stats}
\end{table}

\subsection{Properties of Interest}
Using the output of the physics simulation together with the inferred component counts we construct a dictionary of 29 aircraft proprieties. We consider the following proprieties to be of particular interest:
\subsubsection{Number of Interferences:} In UAV design, an interference is a condition where components in the CAD model overlap (such as a propeller cutting through the fuselage). It is possible to predict the number of interferences from the design using geometric reasoning.

\subsubsection{Number of propellers:} Most designs in the AircraftVerse dataset have an even number of propellers, usually from 2 to 8. We note that the number of propellers can be directly inferred from the design tree, which has relevance in the surrogate modeling task.
\subsubsection{Mass:} The total mass of the UAV design in kilograms.
\subsubsection{Hover time:} The total time (in seconds) that the design is able to hover according to the physics simulation.
\subsubsection{Maximum distance:} The total distance (in meters) that the aircraft is able to fly according to the physics simulation.
\subsubsection{Maximum speed:} The maximum speed (in meters per second) that the aircraft can achieve according to the physics simulation.

% \section{Our Approach}
\section{The Aircraft Generator Pipeline}
\subsection{CodeT5+ Model Description}
Recent success in large language modeling has led to impressive performance on many natural language understanding tasks without the need for task-specific fine-tuning \cite{bloom,llama}. Similar models trained on huge, publicly available code datasets (e.g., GitHub) show the ability to generate, understand, and autocomplete code \cite{nijkamp2023codegen2,codet5p}. Due to JSON's similarity to code, we chose to build AGENT using the recent code LLM called CodeT5+ \cite{codet5p}.

CodeT5+ is an encoder-decoder transformer with an architecture inherited from the T5 model \cite{t5}. CodeT5+ achieves state-of-the-art performance by training a single model on a suite of tasks. It can be trained in encoder-only, decoder-only, and encoder-decoder modes. The decoder is trained using a causal language modeling loss. In addition, the encoder is used to condition generation using natural language prompts (i.e., prompt-based generation). Furthermore, the model is simultaneously trained using a masked language modeling loss, which allows the model to better capture low-level details of the code. A key component of CodeT5+'s success is that it is trained simultaneously on a diverse set of tasks.

Inspired by the approach of CodeT5+, we design a curriculum of learning tasks that encourages the model to learn richer representations of the data while also imbuing a single model with a variety of controllable behaviors. We train our model in encoder-decoder mode using two input sequences: $encS$ and $decS$. 
For example $decS$ is a sequence of length $N$, i.e. $decS = t_0,t_1,\dots,t_{N-1}$ for some tokens $t_i$. During training the input to the decoder is the first $N-1$ tokens $t_0,t_1,\dots,t_{N-2}$. The training targets for the softmax layer of the language model are tokens of $decS$ shifted to the right: $t_1,t_1,\dots,t_{N-1}$. In this way, the model is trained to predict the next token for each input token using a cross entropy loss. This is known as causal language modeling \cite{codet5p}. The sequence $encS$ is fed into the encoder which generates an embedding that is then cross attended with an intermediate representation of the decoder before making the next token prediction. With this approach, we can generate a varied curriculum of tasks for the model by simply defining ($encS$, $decS$) pairs of sequences. In general we think of the encoder as providing a command (possibly with additional context) to the decoder which then generates a relevant sequence.

The CodeT5+ model comes in various sizes, ranging from 220 million to 16 billion parameters. In this work, we use the smallest member of the family, as it is trainable and deployable using a single A100 GPU. This allows AGENT to be a practical tool for human designers, operable on a desktop workstation.

\begin{table*}[h]
\centering
\begin{tabular}{cc|ccc}
\toprule
Components? & Masking? & Hover Time & Max Speed & Max Distance \\
\midrule
\textcolor{red}{\ding{55}} & \textcolor{red}{\ding{55}} & 0.888 & 0.927 & 0.928 \\
\textcolor{green}{\checkmark} & \textcolor{red}{\ding{55}} & 0.893 & \textbf{0.944} & \textbf{0.944} \\
\textcolor{red}{\ding{55}} & \textcolor{green}{\checkmark} & 0.907 & \textbf{0.944} & \textbf{0.944} \\
\textcolor{green}{\checkmark} & \textcolor{green}{\checkmark} & \textbf{0.908} & 0.941 & 0.942 \\
\bottomrule
\end{tabular}
\caption{Binary Classification Results. We compute the test set accuracy on successful and failed designs separately. We then report the harmonic mean between the two.}

\label{table:classification_results}
\end{table*}
\begin{table*}[h]
\centering
\begin{tabular}{cc|ccc|ccc}
\toprule
Components? & Masking? & \# Interferences & \# Propellers & Mass & Hover Time & Max Speed & Max Distance \\
\midrule
\textcolor{red}{\ding{55}} & \textcolor{red}{\ding{55}} & 0.943 & 0.980 & 0.989 & 0.356 & 0.161 & 0.269 \\
\textcolor{green}{\checkmark} & \textcolor{red}{\ding{55}} & \textbf{0.957} & 0.980 & 0.980 & 0.289 & 0.027 & -1.152 \\
\textcolor{red}{\ding{55}} & \textcolor{green}{\checkmark} & 0.923 & \textbf{0.995} & 0.989 & -1.948 & 0.059 & \textbf{0.331} \\
\textcolor{green}{\checkmark} & \textcolor{green}{\checkmark} & 0.938 & 0.980 & \textbf{0.992} & \textbf{0.418} & \textbf{0.427} & 0.245 \\
\bottomrule
\end{tabular}
\caption{Surrogate model results. We report the $R^2$ between predicted and ground-truth values.}
\label{table:surrogate_results}
\end{table*}

\subsection{Domain-Specific Curriculum for Aircraft Generation and Surrogate Modeling}

\paragraph{(1) Learning the Component Corpus.}
For the component dataset, we consider three tasks, as shown in the top row of Figure \ref{fig:comp_tasks}. First, we address a component generation task. The encoder is given the command \texttt{<|generate\_component|>} along with the name of a component in the database. The model's target is to learn a JSON dictionary representing the component properties. We also consider the reverse operation: given the component properties, the model must predict the component's string name. Lastly, we tackle a mask-filling task. The encoder receives the command \texttt{fill\_masks} along with a masked version of the component properties. The decoder must then output a JSON dictionary with the correct values to fill in the mask.

We note that the component names, which occur in the design trees, are represented by the same tokens. Although the component information is never directly linked to the designs, the token embeddings are shared across tasks. Therefore, by training a single model on both the component database and design trees, the model learns a latent connection between them.

\paragraph{(2) Conditional Design Generation Task.}
For the dataset of UAV designs, we adopt a similar strategy, as illustrated in the bottom row of Figure \ref{fig:comp_tasks}. First, we consider a conditional design generation task. The encoder is given the command \texttt{<|generate\_design|>} along with the JSON dictionary from \texttt{output.json}. The target for the decoder is the corresponding design tree in the JSON format. During training, we randomly condition on a subset of the output dictionary. This flexibility allows us to condition on any number of properties during inference without needing to provide the entire output dictionary. In the most extreme case, where the encoder is given only the generation command, this becomes equivalent to traditional language modeling.

\paragraph{(3) Design Tree Masking Task.} Like with the component database, we add a masking task to the curriculum. For this task, we leverage the structure of the design tree. We first consider masking only the leaves of the design tree. This corresponds to the fine-grained numerical properties of the design (e.g., the length of the fuselage). We then consider masking entire subtrees. This enables the model to predict large components of the design (e.g., the fuselage) conditioned on the rest of the design (e.g., the hub). This capability is especially useful, enabling a human operator to specify certain design requirements while the model fills in the remaining design details. In the most extreme case, the operator could input only a design skeleton, providing the model with another method to generate entire designs.

\paragraph{(4) Surrogate Modeling Task.} To round out AGENT's capabilities as a co-designer, we introduce a design evaluation task. This task enables AGENT to act as a surrogate model of the physics simulator. The encoder receives the input command \texttt{<|evaluate\_design|>} along with the JSON string representation of a design tree. The target for the decoder is the JSON data from \texttt{output.json}, complemented by component counts inferred from the design tree.

\subsection{Training Details}
For all our experiments, we start from a pretrained CodeT5+ model with 220 million parameters. During training, we fine-tune all the model parameters. We observed better performance by fine-tuning all weights rather than freezing certain components, such as the decoder, as suggested in \citet{codet5p}. We train and evaluate the model using 16-bit floating-point precision. The model was trained using the AdamW optimizer with a weight decay of 0.1. We experimented with lower weight decay but found it led to reduced diversity in the generated output. We trained for 50,000 steps with a batch size of 50 on a single NVIDIA A100 GPU. We set the initial learning rate to $10^{-4}$. From the 40,000th step, we exponentially decay the learning rate until it reaches a minimum value of $10^{-6}$. We applied gradient clipping with a maximum norm of 1. We balanced each batch to include 10\% of samples from the component data, 45\% from successful designs, and 45\% from failed designs. The tasks in our curriculum were chosen randomly for each batch element.

\subsection{Inference Details}
Like most LLMs, AGENT uses autoregressive beam search decoding during test-time inference. We apply it in two ways: greedy and sampling. In the greedy setting, AGENT selects the highest probability output path, considering a beam size of $B$ potential paths at each auto regressive step. In sampling mode AGENT draws $B$ samples from the categorical distribution predicted by the language model head. This allows AGENT to generate multiple different outputs for the same prompt.

\begin{table*}[h]
\centering
\begin{tabular}{ccc|ccc}
\toprule
\multicolumn{3}{c|}{Prompt}                & \multicolumn{3}{c}{Average Result from Simulator} \\
\toprule
Hover Time (s) & Max Distance (m) & \# Propellers & Hover Time (s)    & Max Distance (m)  & \# Propellers   \\
\midrule
0          & -            & -             & 0             & 0               & 4               \\
100        & -            & -             & 197.07        & 3924.2          & 6               \\
250        & -            & -             & 201.4         & 3744.8          & 6               \\
\midrule
-          & 0            & -             & 0             & 0               & 4               \\
-          & 2000         & -             & 139.8         & 2830.0          & 6               \\
-          & 3000         & -             & 118.62 & 2887.4          & 6               \\
\midrule
100        & -            & 4             & 67.7          & 1139.6          & 4               \\
100        & -            & 6             & 157.1         & 2520.0          & 6               \\
100        & 3000         & 6             & 172.0         & 2970.2          & 6               \\
\bottomrule
\end{tabular}
\caption{Conditional Generation Results. On the left-hand side we show the prompt to the model. On the right, we report the average value for the metrics after running 5 generated designs (from each prompt) through the physics simulator.}
\label{table:cond_gen}
\end{table*}

\section{Experiments and Results}
\subsubsection{Surrogate Model Results}
With the \texttt{evaluate\_design} command, we can use AGENT as a surrogate model for the physical simulation underlying the AircraftVerse dataset. This capability is especially advantageous since a single forward-pass through AGENT is significantly faster than executing the flight dynamics model (see Table \ref{table:runtime}). To test AGENT's ability to act as a surrogate model, we execute the \texttt{evaluate\_design} command on the 510 held-aside test designs. The predictions are generated using greedy decoding with a beam size of 10. We conduct an ablation study on the content of the training set. We consider the baseline model to be the one trained solely for generating and evaluating designs. Subsequently, we explore the addition of the component database (all tasks) and/or the masking of the design tree (both leaves and subtrees).

\paragraph{Determination of a Successful Design.} We first construct a binary classification to assess the model's ability to detect successful versus failed designs. We consider a classification to be correct if the model predicts a value greater than 0 for a successful design and 0 for failed designs. We compute accuracy separately for both successful and failed designs on a test set of unseen designs. We then report the harmonic mean of these values (see Table \ref{table:classification_results}). For all ablation scenarios, AGENT demonstrates high accuracy in detecting whether a design will be successful. This outcome underscores AGENT's capability to serve as a surrogate model for the simulator, highlighting a valuable feature of our approach. It enables a human designer to quickly determine if a candidate design would be able to fly, greatly accelerating the design process.

\paragraph{Design Metric Prediction.} Beyond the classification task, we evaluate the accuracy of AGENT's predictions for specific metrics by computing the coefficient of determination ($R^2$) between its predictions and the ground truth. Table \ref{table:surrogate_results} presents the results of this experiment. For this experiment, we use only the test set of successful designs, which considerably increases the task's difficulty. Initially, we focus on predicting a design's physical properties. These properties include the number of interferences between components, the number of propellers, and the overall mass.

While accurately predicting the number of interferences necessitates implicit geometric understanding, inferring other physical properties from the design tree is more straightforward. All model configurations demonstrate a strong correlation with the ground truth values. This confirms that AGENT's surrogate modeling mode can effectively recover properties directly inferable from the design.

Next, we consider predicting the following flight dynamics of the model: the maximum hover time, the maximum flight distance, and the maximum speed. This is a more challenging task as it requires an understanding of all the interacting components of a design, including the structure and the motor-propeller-battery combination. Furthermore, it necessitates an implicit understanding of the physical dynamics of UAV flight. As a result of this complexity, we observe a weaker correlation than in the previously described simpler experiment. Interestingly, we observe a complex interaction when employing different training tasks. For example, incorporating the component database tasks results in less accurate hover time predictions. However, when we add the design masking tasks to the training regimen, the model achieves optimal performance. This supports our approach of employing a diverse set of training tasks within a single model generally leads to enhanced performance. Additionally, we emphasize that the challenge of this task was amplified by focusing solely on predicting the results of the ``successful" designs, given that our models have already proven adept at filtering out ``failed'' designs.

\subsubsection{Conditional Generation Results}
Next, we evaluate AGENT's ability to perform conditional generation of UAV designs. We create a set of prompts and generate 5 designs using sampling mode with a beam size of 10. We consider conditioning on hover time, maximum flight distance, and number of propellers. After generating a set of designs for each prompt, we run them through the physics simulator of the AircraftVerse dataset. This allows us to accurately detect how closely the generated designs align with the desired prompt. We report the prompts and results of this experiment in Table \ref{table:cond_gen}.

First, we note that AGENT consistently generates designs with the specified number of propellers. This is a valuable feature in itself, as it would allow a human designer to enforce design constraints imposed by component availability (e.g., when only four propellers are available). When conditioning on flight dynamic properties, the alignment between the generated designs and the simulator is less precise, with the exception of conditioning on a hover time or maximum distance of 0 (e.g. a failed design). In this case simulator results exactly match the prompt, which isn't surprising given the results in Table \ref{table:classification_results}.

\begin{table}[]
\begin{tabular}{c|c|c}
\toprule
GPU           &  GPU RAM Usage (\%) & Run Time   (s) \\
\midrule
%3080TI (12GB) & 19.6            & TBD        \\
 A6000 (40GB)  & 5.4             & 12         \\
A100 (80GB)   & 3.2             & 11         \\
Simulator     & 0               & $\sim$120 \\
\bottomrule
\end{tabular}
\caption{Runtime Comparison}
\label{table:runtime}
\end{table}

\begin{figure}[]
    \centering    \includegraphics[width=0.9\columnwidth]{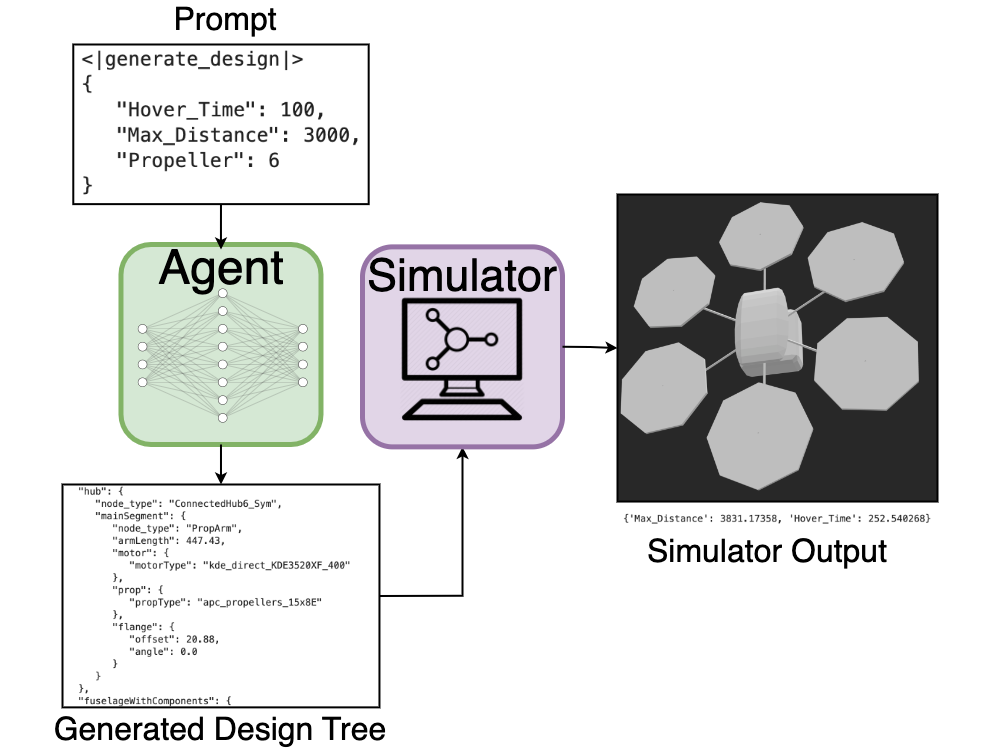}
    \caption{Example generation using multiple prompts. We ran the generated design into the physics simulator to verify the design. In practice, this would be optional.}
    \label{fig:example_gen}
\end{figure}

When conditioning for successful designs, we see that the generated designs do not exactly match the simulators output, however the results are always within the same order of magnitude of the prompt. We are also able to generate designs conditioned on multiple parameters simultaneously. We show an example of this in Figure \ref{fig:example_gen}. This is an especially useful feature of AGENT. It would allow a human designer to quickly generate a design that is flyable, which could then be iterated on (with  the continued help of AGENT).
\section{Path to Deployment}
\subsubsection{Runtime Analysis}
To evaluate the feasibility of deploying AGENT, we first conduct a runtime analysis.
For a suite of GPUs, we generate a sequence of 512 tokens, the maximum possible length. Firstly, we observe that AGENT is memory-efficient, utilizing only about 2.5 gigabytes of GPU memory. This efficiency stems from our use of the smallest member of the CodeT5+ family, which operates with 16-bit floating precision. If available, more memory could accommodate a larger beam size or facilitate the parallel generation of multiple designs (which we did to generate Table \ref{table:cond_gen}). Moreover, our code is adaptable for training and deploying larger variants of the CodeT5+ family, which we leave for future work.

Furthermore, we observe that AGENT operates an order of magnitude faster than the AircraftVerse dataset's physics simulator. This speed advantage allows a human designer to iterate on designs significantly faster than with traditional workflows. For instance, within seconds, a designer can determine the viability of a candidate design with high precision. Moreover, in the time it takes to run a single iteration of the simulator, a designer could complete multiple design loop iterations using AGENT. Overall the runtime performance of AGENT makes it well suited to deployment.

\subsubsection{Human-AI Collaboration}
%The UAV components in AircraftVerse dataset are commodity parts which are widely available. Therefore designs generated by AGENT can quickly be turned into physical prototypes. 

A challenge that we are currently working on is the development of a user-friendly UI front-end tailored for human designers. While AGENT's current implementation is efficient and is already a useful design tool, it necessitates the execution of Python code using JSON prompts. We highlight that the model can be readily adapted to accept natural language prompts through chatbot platforms like GPT \cite{openai2023gpt4} or LLaMA \cite{llama}. This adaptation would facilitate a dynamic human-in-the-loop workflow, enabling continuous design updates and evaluations.

\subsubsection{Simulator-AI Integration} Additionally, integrating the physics simulator directly into AGENT presents a promising avenue. By doing so, AGENT could autonomously update its surrogate model, leading to a reduction in human intervention for specific aspects of the design process. Such an integration would not only streamline the workflow but also ensure higher quality design generation. By granting AGENT direct access to the physics simulator, the system can make real-time adjustments and refinements, ensuring improved design outcomes.

\section{Conclusion}
In this paper, we presented AGENT, a design and evaluation tool for UAVs. Leveraging the recent AircraftVerse dataset and state-of-the-art CodeT5+ LLM, we trained a unified model directly on JSON files. AGENT is capable of conditionally generating designs while simultaneously serving as a surrogate for the physics simulator of the AircraftVerse dataset. We validated AGENT's rich set of capabilities through a series of experiments. Lastly, we considered the feasibility and remaining challenges for deploying AGENT as a tool for human designers. Overall, we believe this work demonstrates a clear path to deployment and we expect to see similar domain-specific LLM co-designers being developed in the near future. 
% We have shown how to develop such a co-designer using our domain-specific curriculum-based approach.

% \section{Ethics Statement}
% We recognize the complex social and  ethical consequences of this work. For example, UAVs are already actively used in law enforcement and surveillance contexts. However, there are many socially beneficial applications of UAVs, as we discussed in the introduction. Additionally, the ideas proposed here could be used to aid the design in many other domains far beyond UAVs. As a result we welcome feedback from the community to ensure that our work is both responsible and beneficial to society, by addressing potential ethical concerns proactively.

% Proactively addressing concerns, we welcome feedback from the community to ensure that our work is both responsible and beneficial, and to help us navigate the complex ethical landscape associated with these technologies.

% We want to promote responsible use and want to address potential ethical concerns proactively.

\section{Acknowledgments}
This project was supported by DARPA under the Symbiotic Design for Cyber-Physical Systems (SDCPS) with contract FA8750-20-C-0002. The views, opinions and/or findings expressed are those of the author and should not be interpreted as representing the official views or policies of the Department of Defense or the U.S. Government.

\bibliography{aaai24}

\end{document}